\documentclass[journal,twoside,web]{ieeecolor}
\usepackage{tmi}
\usepackage{cite}
\usepackage{amsmath,amssymb,amsfonts}
\usepackage{algorithmic}
\usepackage{graphicx}
\usepackage{textcomp}
\usepackage{multirow,tabularx}
\usepackage{subfig}

\def\BibTeX{{\rm B\kern-.05em{\sc i\kern-.025em b}\kern-.08em
    T\kern-.1667em\lower.7ex\hbox{E}\kern-.125emX}}
\begin{document}
\title{Surgical Activity Recognition Using Learned Spatial Temporal Graph Representations of Surgical Tools}
\author{Duygu Sarikaya and Pierre Jannin \thanks{``This work was supported by French state funds managed within the Investissements d'Avenir program by BPI France (project CONDOR).''}  \thanks{D. Sarikaya was with University of Rennes 1, INSERM, LTSI - UMR 1099, Rennes 35000 France, She is now with Department of Computer Engineering, Faculty of Engineering, Gazi University, 06570 Ankara, Turkey (e-mail: duygusarikaya@gazi.edu.tr).}  \thanks{ P. Jannin is with University of Rennes 1, INSERM, LTSI - UMR 1099, Rennes 35000 France (e-mail: pierre.jannine@univ-rennes1.fr).}}

\maketitle

\begin{abstract}

Modeling and recognition of surgical activities poses an interesting research problem. Although a number of recent works studied automatic recognition of surgical activities, generalizability of these works across different tasks and different datasets remains a challenge. We introduce a modality that is robust to scene variation, and that is able to infer part information such as orientational and relative spatial relationships. The proposed modality is based on spatial temporal graph representations of surgical tools in videos, for surgical activity recognition. To explore its effectiveness, we model and recognize surgical gestures with the proposed modality. We construct spatial graphs connecting the joint pose estimations of surgical tools. Then, we connect each joint to the corresponding joint in the consecutive frames forming inter-frame edges representing the trajectory of the joint over time. We then learn hierarchical spatial temporal graph representations using Spatial Temporal Graph Convolutional Networks (ST-GCN). 
Our experiments show that learned spatial temporal graph representations perform well in surgical gesture recognition even when used individually. We experiment with the \textit{Suturing} task of the JIGSAWS dataset where the chance baseline for gesture recognition is $10$\%. Our results demonstrate $68$\% average accuracy which suggests a significant improvement. Learned hierarchical spatial temporal graph representations can be used either individually, in cascades or as a complementary modality in surgical activity recognition, therefore provide a benchmark for future studies. To our knowledge, our paper is the first to use spatial temporal graph representations of surgical tools, and pose-based skeleton representations in general, for surgical activity recognition.

\end{abstract}


\section{Introduction}

Modeling and recognition of surgical activities poses an interesting research problem as the need for assistance and guidance through automation is addressed by the community. Although a number of recent works studied automatic recognition of surgical activities, generalizability of these works remain a challenge. Moreover, the need for representations with greater expressive power, such as graphs, that we can use not only to recognize surgical activities but also to bridge the gap between recognition and control in autonomous systems is growing. 

\begin{figure*}[!t] 
\centering
 {\includegraphics[width = 7.2in]{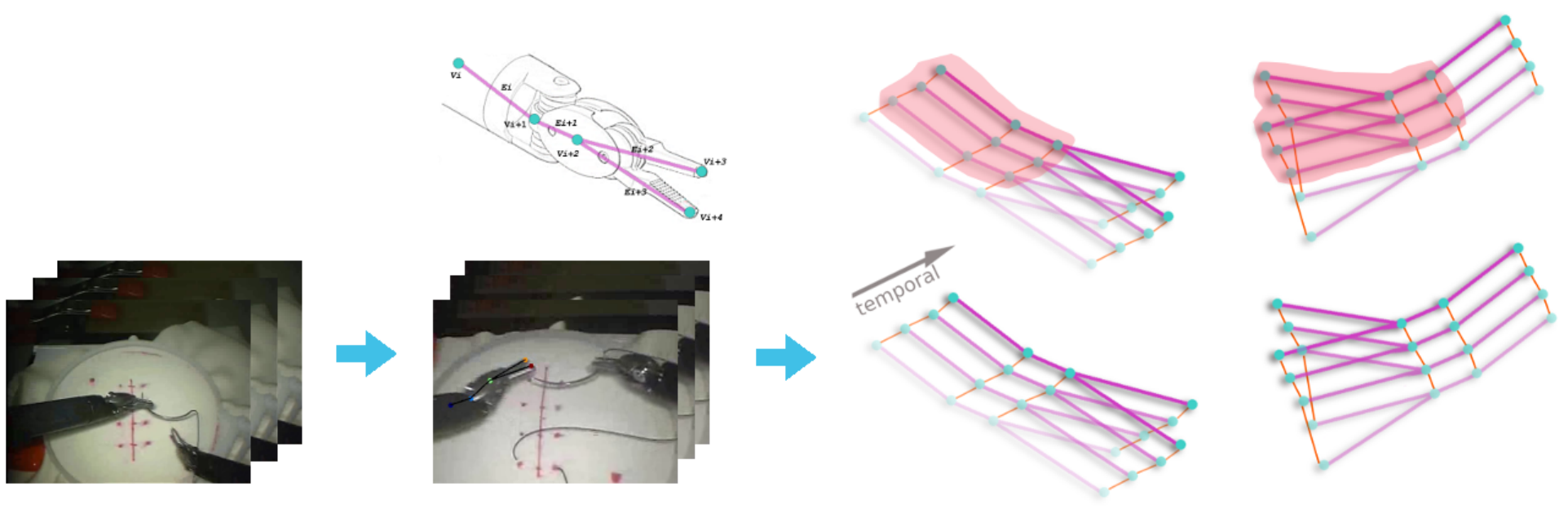}}
\caption{For each video segment, we construct an undirected spatial temporal graph of the joints over temporal sequences of frames. Then through multiple layers of spatial-temporal graph convolution (ST-GCN), we learn hierarchical spatial temporal representations of surgical activities. (Image Reference (drawing of the grasper): \cite{Kim2014}) }
\label{fig:overview}
\end{figure*}

Frame based image cues have been widely used for recognition of surgical activities. Although these studies have been tremendously successful in terms of high accuracy, a major setback is that their performances are limited to the dataset they are modeled on, and they are often prone to overfitting. The generalizability across different tasks and datasets remains a challenge. For example, placing a \textit{Tie Knot} might occur during a task of \textit{Suturing on Tissue} and also during the more specific and challenging task of \textit{Urethrovesical Anastomosis (UVA)} that involves stitching and reconnecting two anatomical structures together \cite{Sarikaya2017}. If we heavily rely on image cues of the surgical scene, representations of these surgical activities vary greatly. 

An example of the limitations of deep neural networks trained on image cues is demonstrated by Mitchell et al. \cite{Mitchell2019}. In this work, the authors discovered that the representations learned were highly dependent on the background instead of the subject being classified, and they were not representative of subject features to be generalized across datasets. Another key challenge in visual recognition is how to accommodate geometric variations in object scale, pose, viewpoint, and part deformation \cite{Dai2017}. In order to tackle this problem, data augmentation is often used. However, the augmentation process involves geometric transformations that are hand-crafted. This prevents generalization to new tasks \cite{Dai2017}. Moreover, Convolutional Neural Networks (CNN) are also shown to have limited capabilities when it comes to inferring part information of objects such as orientational and relative spatial relationships \cite{Sabour2017}. These limitations affect the generalizability of tasks to varying degrees depending on the task's nature. For example, automatic pose estimation of surgical tools, which are rigid objects, are one of the better solved problems in surgical video understanding  \cite{Bouget2017} as the geometric variability of the surgical tools are limited. However, it is more difficult to generalize gesture recognition across different datasets as the spatial and temporal dynamics of a gesture are not readily formulated. Kinematic data captured from the surgeon and patient side manipulators (robotic arms), which are operating out of the patient's body, are also limited as they don't provide sensitive enough kinematic information relating the surgical tool end effectors which are operating on the surgical scene.

In order to address the limitations mentioned and overcome the challenge of generalizability across different tasks and different datasets, we need to define more generic and also more sparse representations of surgical activities that are robust to scene variation. Scene invariant approaches using various modalities such as optical-flow, depth, and skeleton representations have been proposed for human activity recognition. Pose estimation based skeleton representations are known to suffer relatively little from the intra-class variances when compared to image cues \cite{yao}. Moreover, using pose-based skeletons sparsely represent the joints and the connections, preventing the dependency on irrelevant cues. They also provide us with part information. Although pose estimation of surgical tools has been studied \cite{Kurmann2017,Du2018,Laina2017,Rieke2015}, pose-based skeleton representations have not been used in surgical activity recognition yet. To our knowledge, our paper is the first to use these representations for surgical activity recognition. 

In this paper, we introduce a modality independent of the scene, that is the background, therefore robust to scene variation, based on spatial temporal graph representations of the surgical tools. We propose to model and recognize surgical activities in surgical videos by first defining the graph representations of the surgical tools over time, and then learning hierarchical spatial temporal representations using Spatial Temporal Graph Convolutional Networks (ST-GCN) \cite{stgcn}. We explore the effectiveness of our model on JIGSAWS dataset \cite{jigsaws} for the \textit{Suturing} task. Figure \ref{fig:overview} shows an overview of spatial temporal graph construction and convolution.

\section{Related Work}

\subsection{Surgical Activity Recognition}

Ahmidi \textit{et al.} \cite{jigsaws_benchmark} did a comparative benchmark study on the recognition of gestures on JIGSAWS dataset. In this study, in order to classify surgical gestures, three main methods are chosen: Bag of Spatio-Temporal Features (BoF), Linear Dynamical System (LDS) \cite{lin2005,lin2006} and a composite Gaussian Mixture Model- Hidden Markov Model: GMM-HMM  \cite{Leong2007,Varadarajan2009,Varadarajan2011}. For BoF, features of both spatially and temporally high texture variations are extracted with Space-Time Interest Points (STIP) \cite{Laptev2005}. This cuboid of features are then combined with additional features such as HOG: histogram of orientation gradients \cite{Dalal2005} and HOF: histogram of optical flow \cite{Dalal2006}. A codebook is created and dimensionality of these visual representations are reduced via clustering. An SVM classifier is trained on videos\rq s histograms of the codebook words. Linear Dynamical System (LDS) on both the image intensities and kinematic data is proposed for the same problem. In this approach, the video frames are modeled as the output of a LDS, then the pairwise distances between the LDS models are measured. Finally, a classifier is trained to predict the class of the gesture frames. In the same work, a composite Gaussian Mixture Model- Hidden Markov Model: GMM-HMM models each gesture as an elementary HMM where each state corresponds to one Gaussian Mixture Model (GMM) on kinematic data \cite{lingling}. Studies that use primarily kinematic data have also been suggested. Ahmidi \textit{et al.} \cite{Ahmidi2013} proposed segmenting and recognizing surgical gestures using similarity metrics on the temporal model of surgical tool motion trajectories defined with descriptive curve coding, which transforms them into a coded string. In addition to the benchmark, more recent works have also proposed using both video and kinematic data \cite{Tao2013,Zappella2013}. Lea \textit{et al.} \cite{Colin2015} proposed to use both modalities to perform segmentation and recognition using object cues and higher-order temporal relationships between action transitions using a variation of Conditional Random Field.

More recently, deep learning architectures \cite{krizhevsky} have been proposed for this open research problem. DiPietro \textit{et al.} \cite{DiPietro2016} proposed using Recurrent Neural Networks (RNN) trained on kinematic data for surgical gesture classification on JIGSAWS dataset. Sarikaya \textit{et al.} \cite{Sarikaya2018} proposed a multi-modal convolutional recurrent neural network architecture whose inputs are video data and motion cues (optical flow \cite{flow}). They proposed to jointly learn surgical tasks and gestures with a multi-task learning approach. Using only optical flow information for surgical activity recognition is also proposed \cite{Sarikaya2019}. The motivation to use motion cues as a joint modality has a similar motivation to ours: generalized models that are more robust to scene variation. However optical flow's performance can be affected by the camera zoom and motion. Lea \textit{et al.} \cite{Colin2016} proposed Temporal Convolutional Network (TCN), that hierarchically captures temporal relationships at low, intermediate, and high-level time-scales for the segmentation of surgical activities, and Convolutional Action Primitives for multimodal time-series data including video data and kinematics to address the same problem. Funke \textit{et al.} \cite{Funke2019} learn 3D convolutional neural networks to learn spatiotemporal features. 3D CNNs applies 3D convolutionals on the 3D temporal representation of the video instead of stacking 2D convolutions at each time frame, however they are known to have problems in training due to the explosion of parameter and they only marginally improve the frame based models on image cues \cite{Wang2013}. 

\subsection{Activity Recognition Using Representations of Joints and Skeletons}

Representations of human body joints and skeletons, and their dynamics have been widely used in open research problems relating video understanding and activity recognition. These Representations of human body joints and skeletons, and their trajectories  are robust to illumination change and scene variation, and they are easy to obtain using depth sensors or pose estimation algorithms \cite{Shotton2011,Cao2017}. Skeletons and joint representations of the hand have also been receiving attention in egocentric cameras and augmented reality systems where the interaction with the real world and the timely response is crucial \cite{Tekin2019,Lepetit2019}. Although, pose estimation of surgical tools has been studied \cite{Kurmann2017,Du2018,Laina2017,Rieke2015}, these representations have not been used for surgical activity recognition yet. 

Human activity recognition methods model the spatial changes of the human joints over a sequence of video frames. Although these methods successfully utilize the orientational and relative spatial relationships of the joints over time to outperform frame based methods, they also have limitations as they often use hand-crafted features or traversal rules \cite{stgcn}.  Xia et al. \cite{Xia2012} use a 3D joint point histogram to represent the human pose, and model the action through a discrete hidden Markov model, while Keceli et al. \cite{Keceli2014} extract human action features based on the angle and displacement information of the skeleton joints \cite{Zhang2019}. Gowayyed et al. \cite{Gowayyed2013} use 3D histogram of oriented displacements (HOD), Hussein et al. \cite{Hussein2013} use covariance matrices of joint trajectories, Wang et al. \cite{Wang2012} use relative positions of joints, and Vemulapalli et al \cite{Vemulapalli2014} use rotations and translations between body parts. Liu et al. \cite{Liu2016} propose spatio-temporal LSTM with trust gates and define a bidirectional tree traversal method to visit joints in a sequence which maintains the adjacency information of the skeletal tree structure. Ke et al. \cite{Ke2017} transform each skeleton sequence into three clips each consisting of several frames for spatial temporal feature learning using deep neural networks, where each clip is generated from one channel of the cylindrical coordinates of the skeleton sequence and each frame represents the temporal information of the entire skeleton sequence, and incorporates one particular spatial relationship between the joints. Spatial Temporal Graph Convolutional Networks (ST-GCN) \cite{stgcn}, on the other hand, applies graph CNNs and learn the part information over spatial and temporal domains implicitly, therefore do not rely on rule based parsing techniques. 

In this paper, we propose to leverage tool skeleton parts information and motion of the surgical tool joints over time for surgical activity recognition. To our knowledge, our paper is the first to use these representations for surgical activity recognition. In this sense, our work differs greatly from related works that use frame based image cues which have limited capabilities of inferring part information of objects such as orientational and relative spatial relationships \cite{Sabour2017}. The modality we propose uses both the spatial relationship information inferred from the joints of a surgical tool and their connectivities, and also the temporal information inferred from the motion of these joints over time across a sequence of frames. Moreover, we propose a modality that is independent of the scene, therefore robust to scene variation.  Through more generic and sparse representations, we can model activities that might occur across different tasks performed under different surgical settings. Related works on the other hand often use frame based image cues, and they are highly dependent on the background of the surgical setting. 

First, we train a CNN to detect the joints of the surgical tools in video frames, and define the surgical tool skeleton. Using the CNN model, we predict the joint coordinates and their confidence scores in a sequence of frames representing each video segment. For each video segment, we construct an undirected spatial temporal graph to form representations of the joints over time. We then learn hierarchical spatial temporal representations using ST-GCN \cite{stgcn} for surgical activity recognition.

\section{Dataset} 

\subsection{JIGSAWS} 

\begin{figure}[!t]
\centering
\subfloat{\includegraphics[width = 1in]{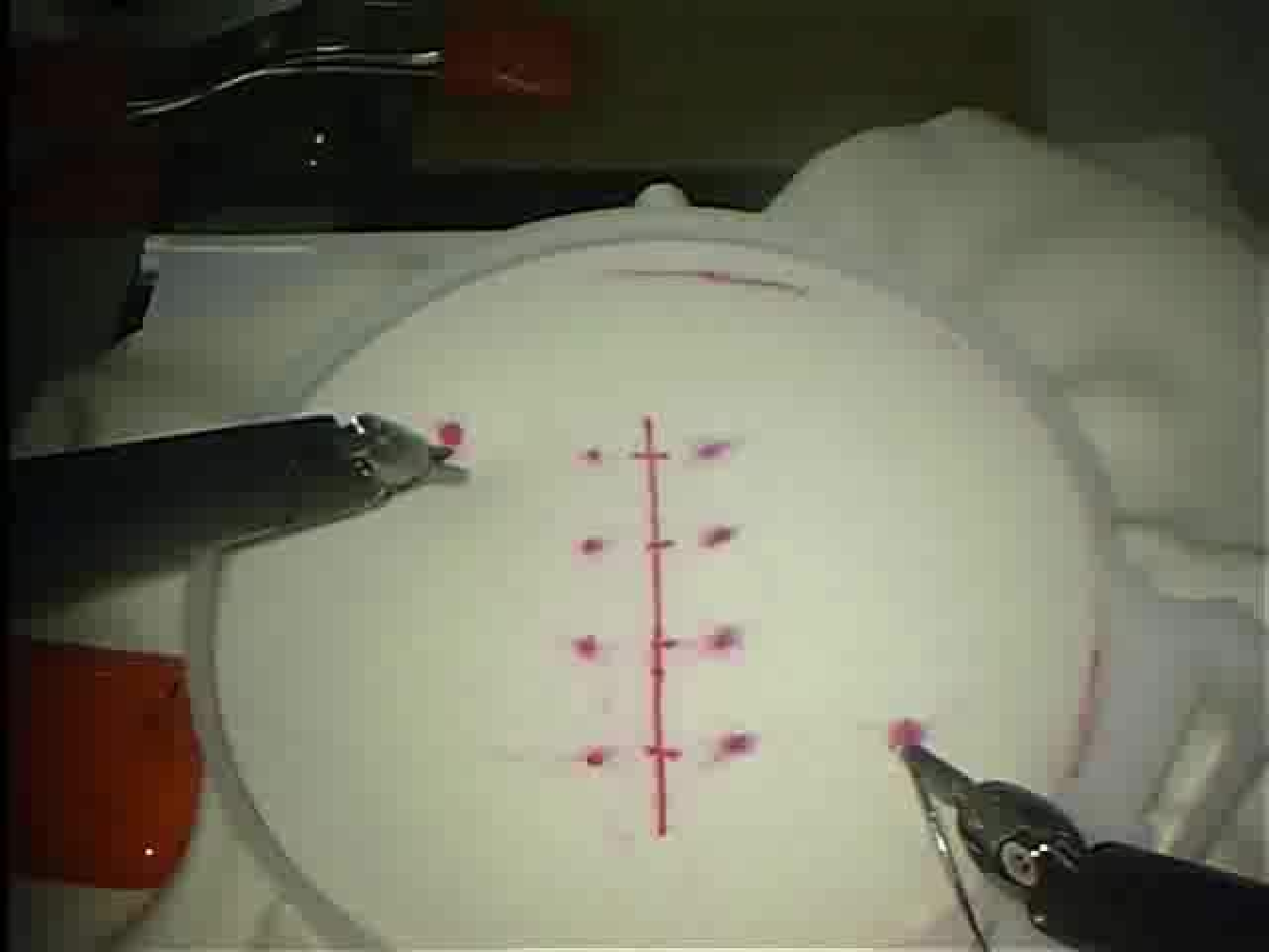}\includegraphics[width = 1in]{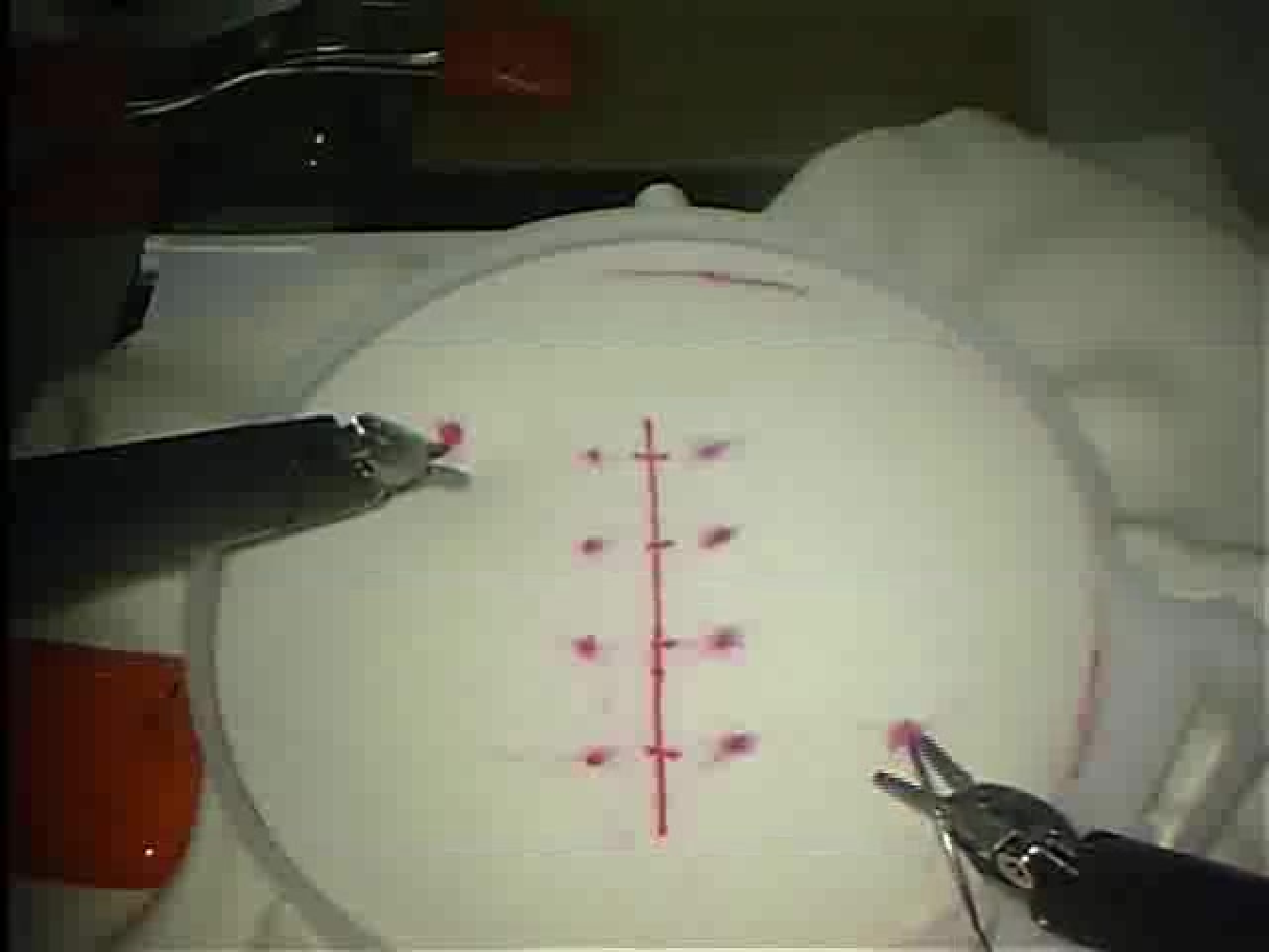}\includegraphics[width = 1in]{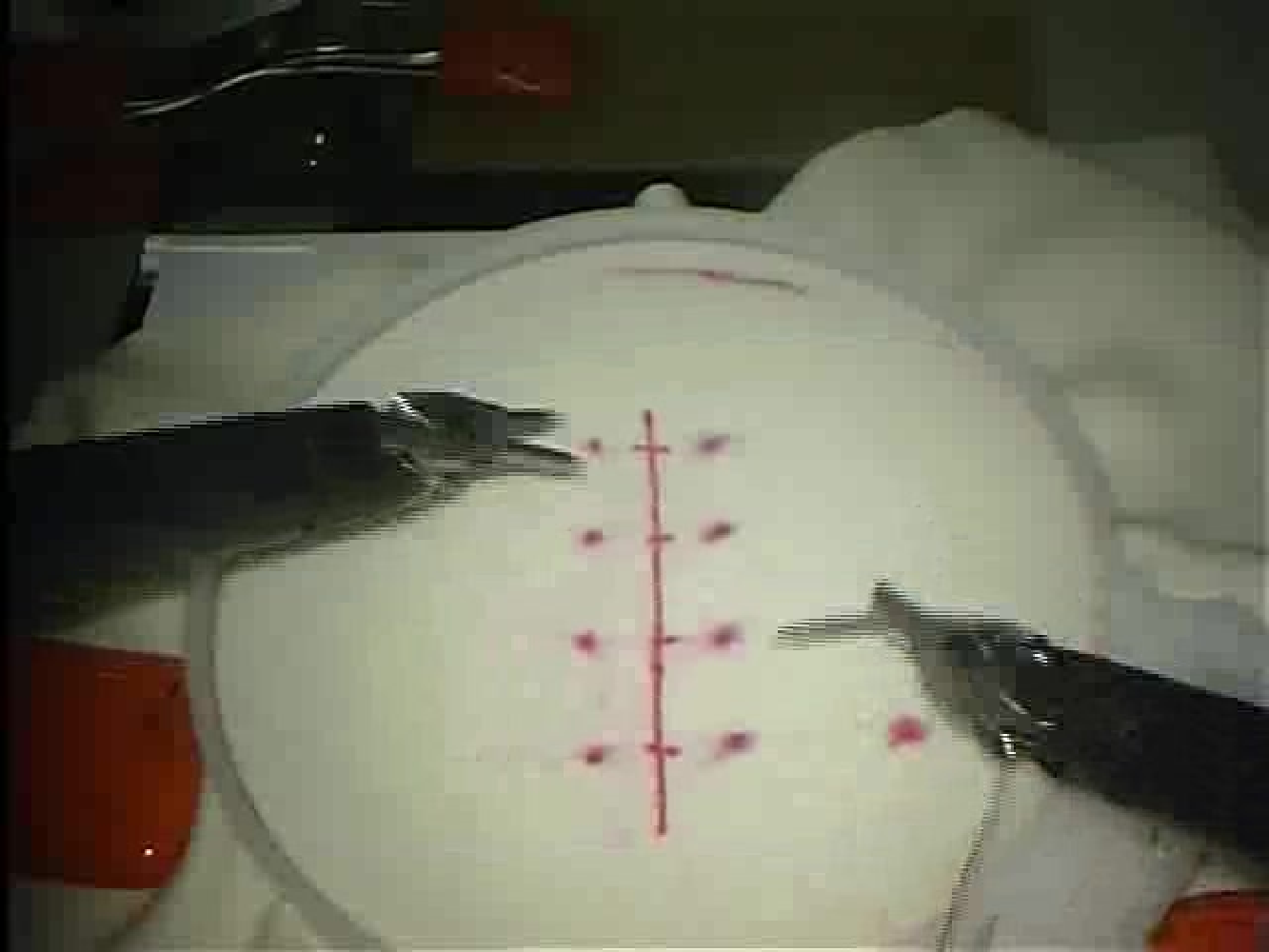}} \\
\includegraphics[width = 1in]{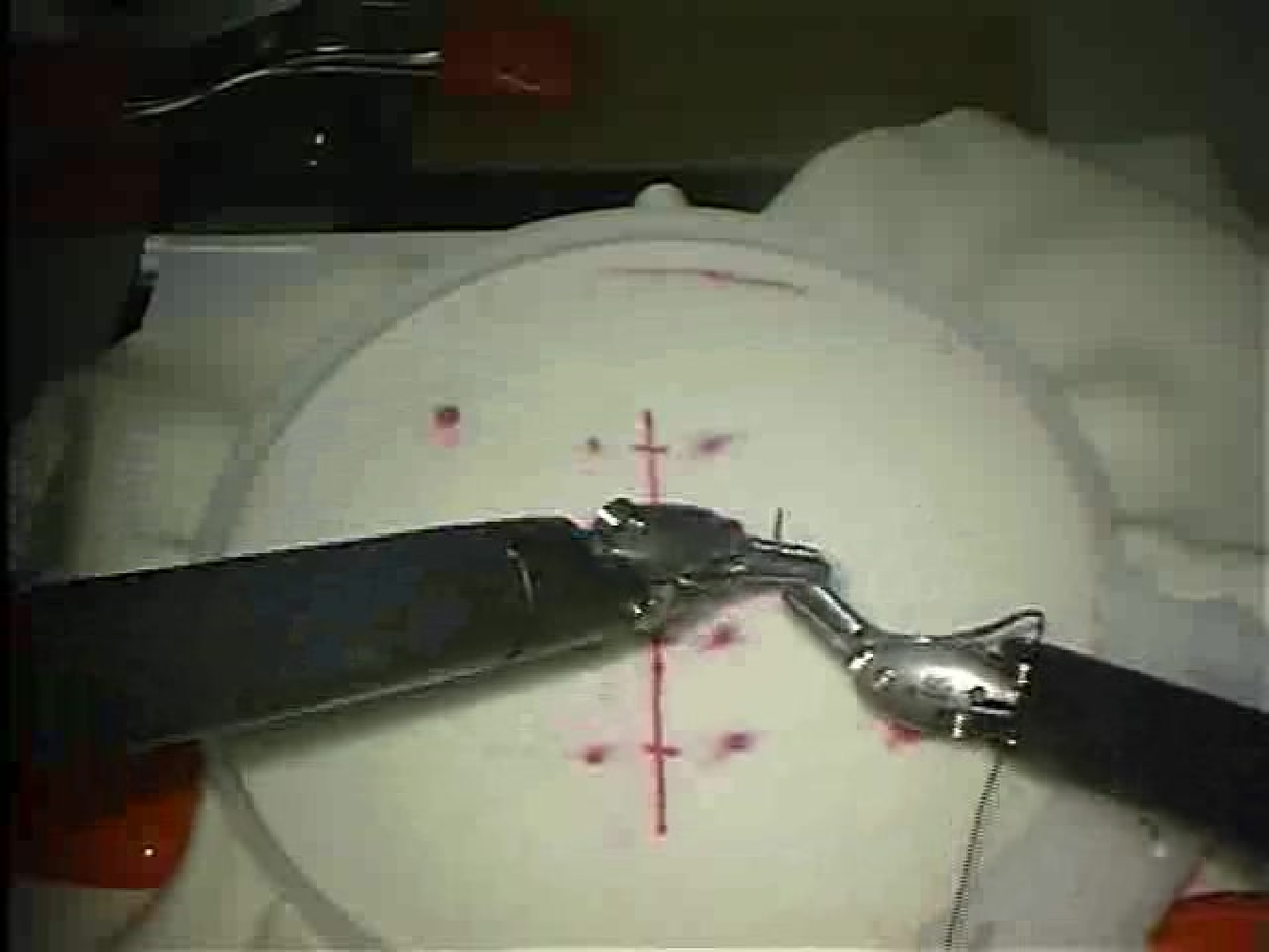}\includegraphics[width = 1in]{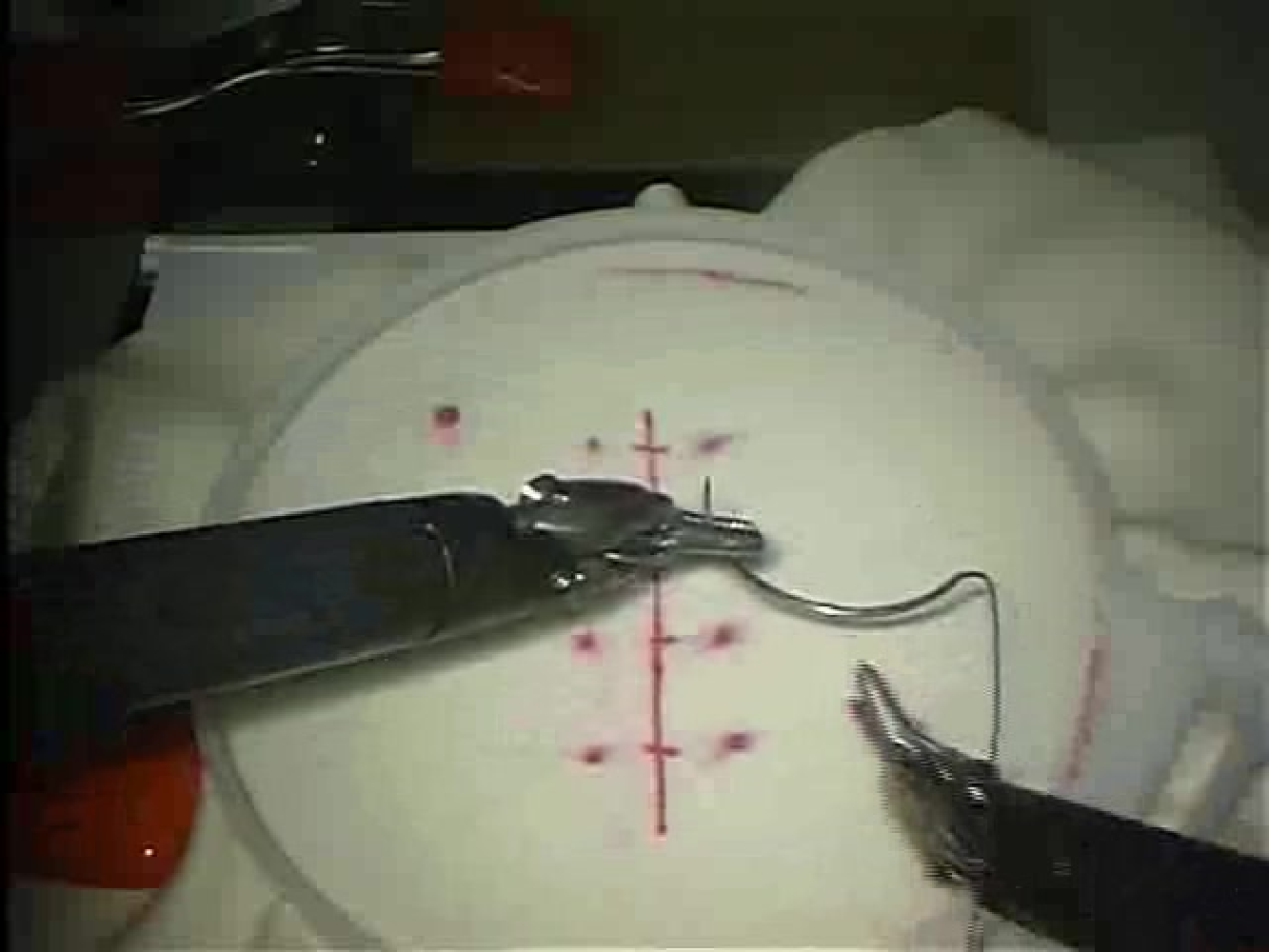}\includegraphics[width = 1in]{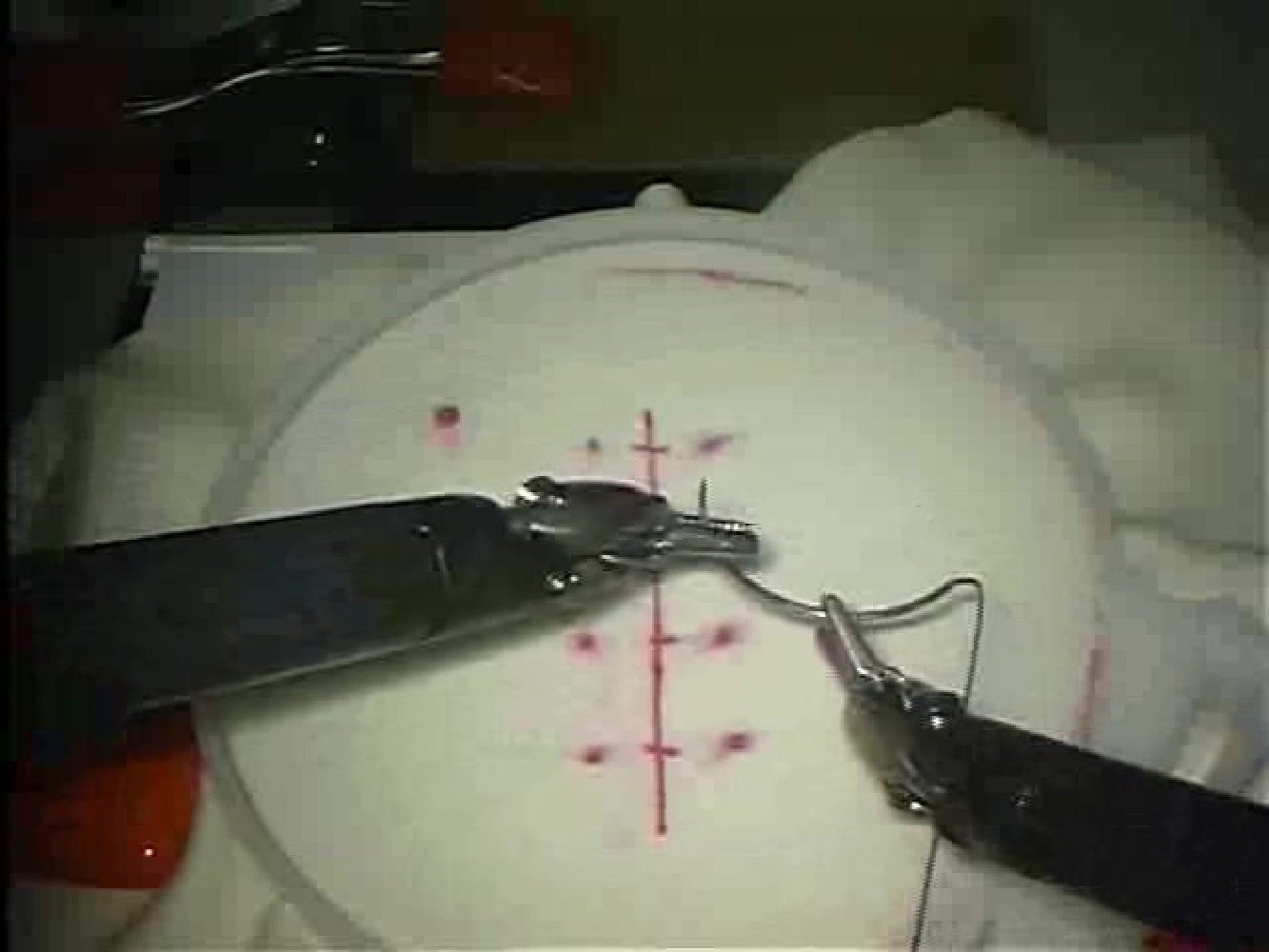} \\
\includegraphics[width = 1in]{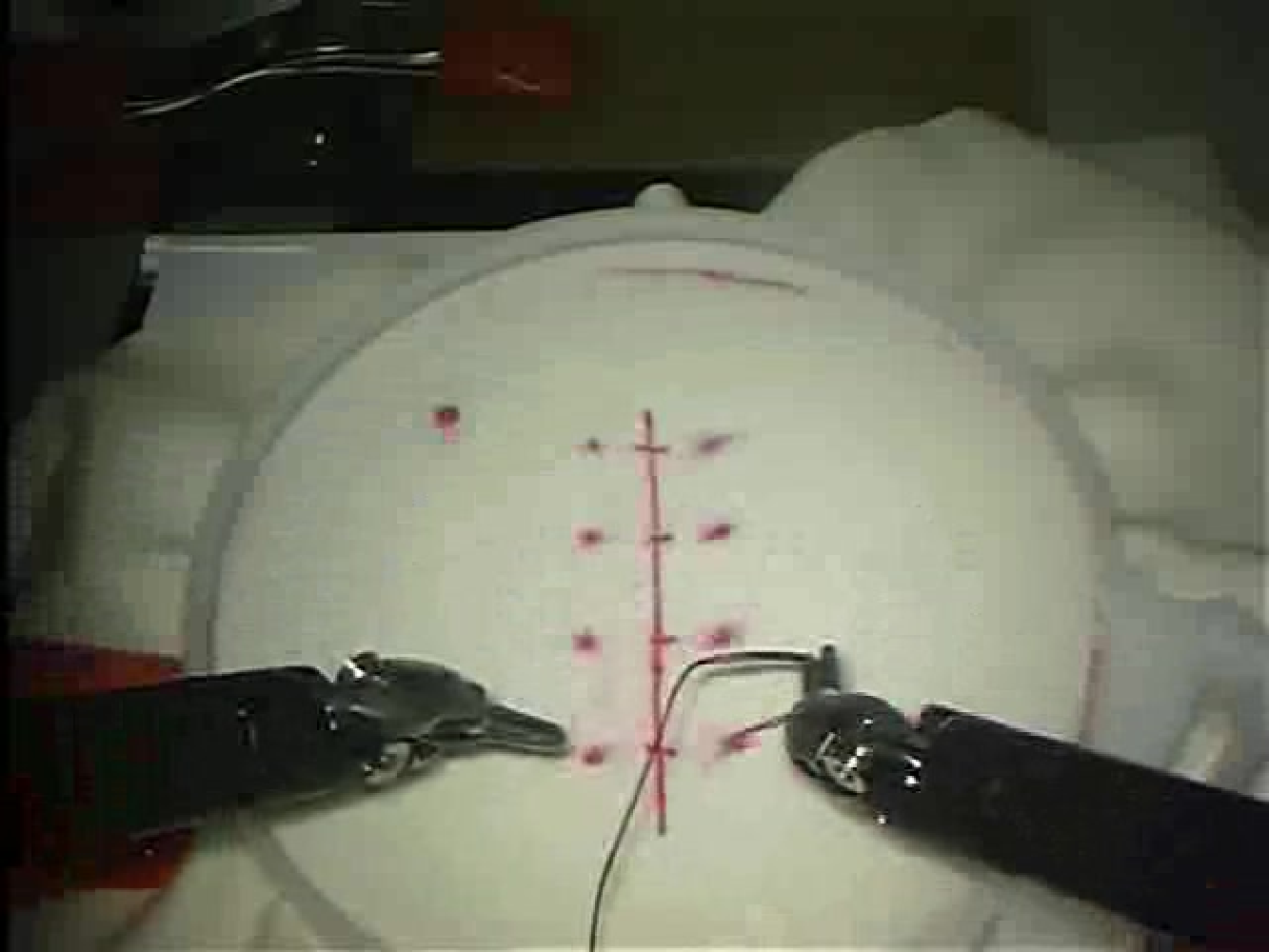}\includegraphics[width = 1in]{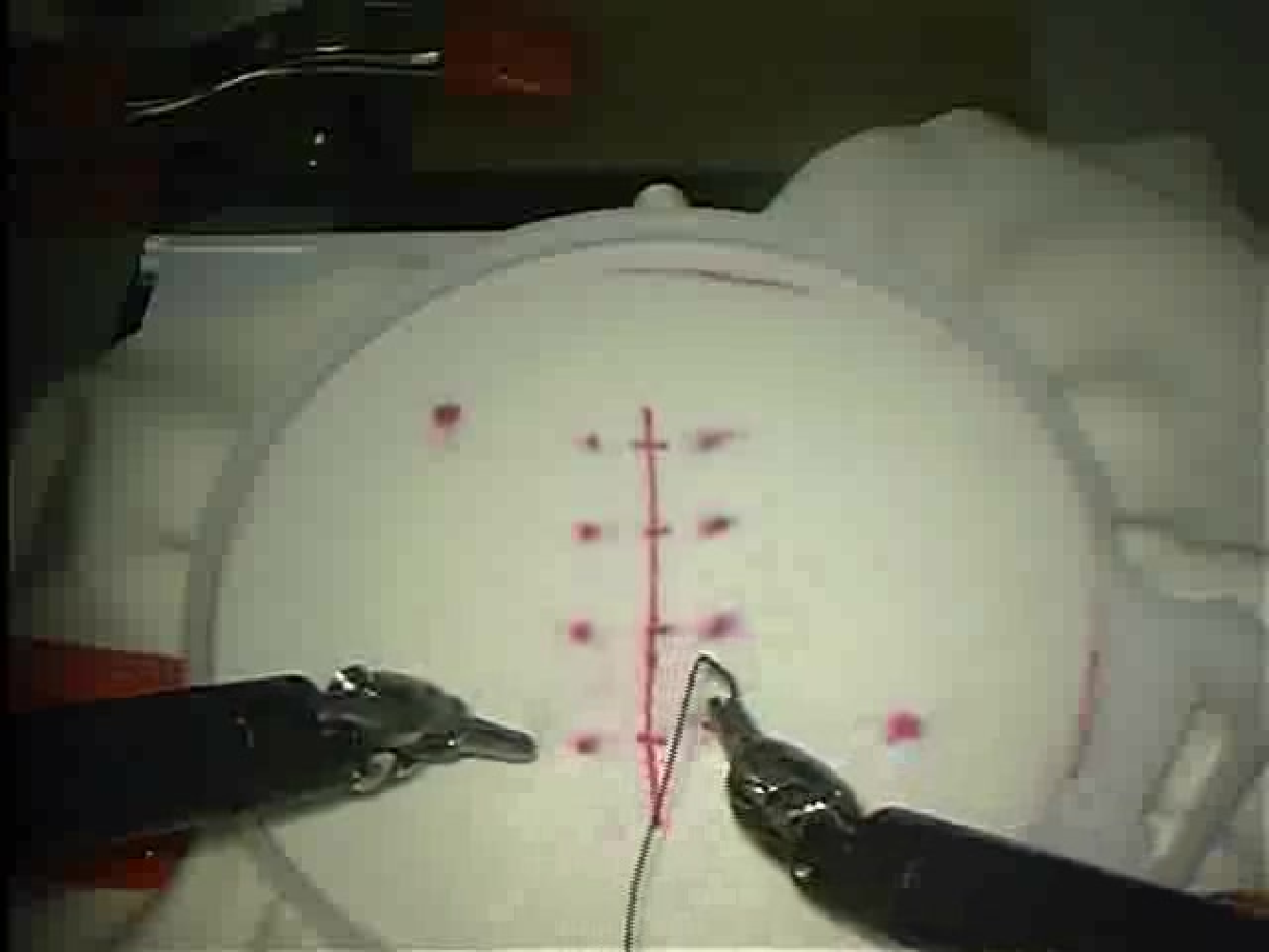}\includegraphics[width = 1in]{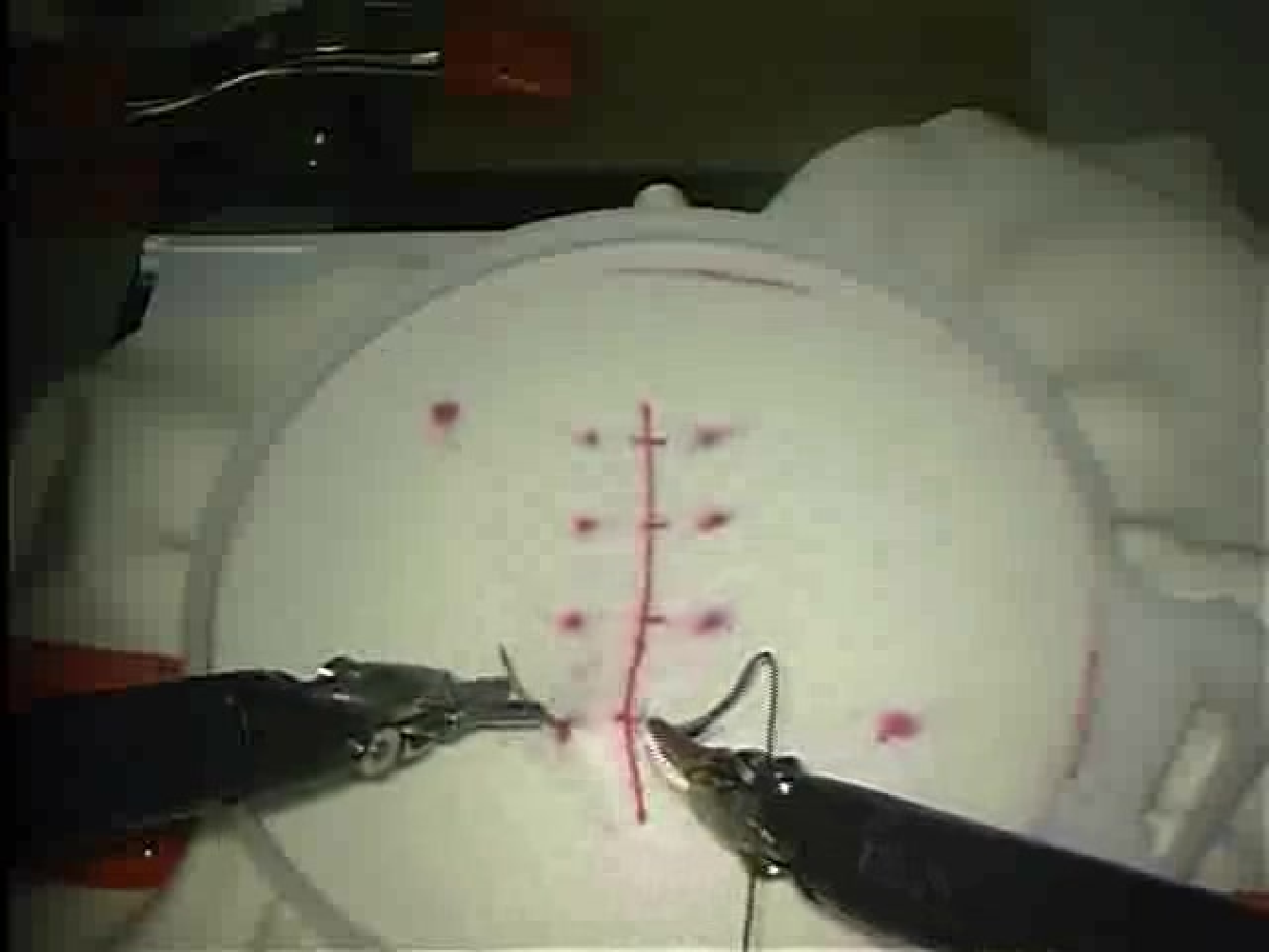} \\
\includegraphics[width = 1in]{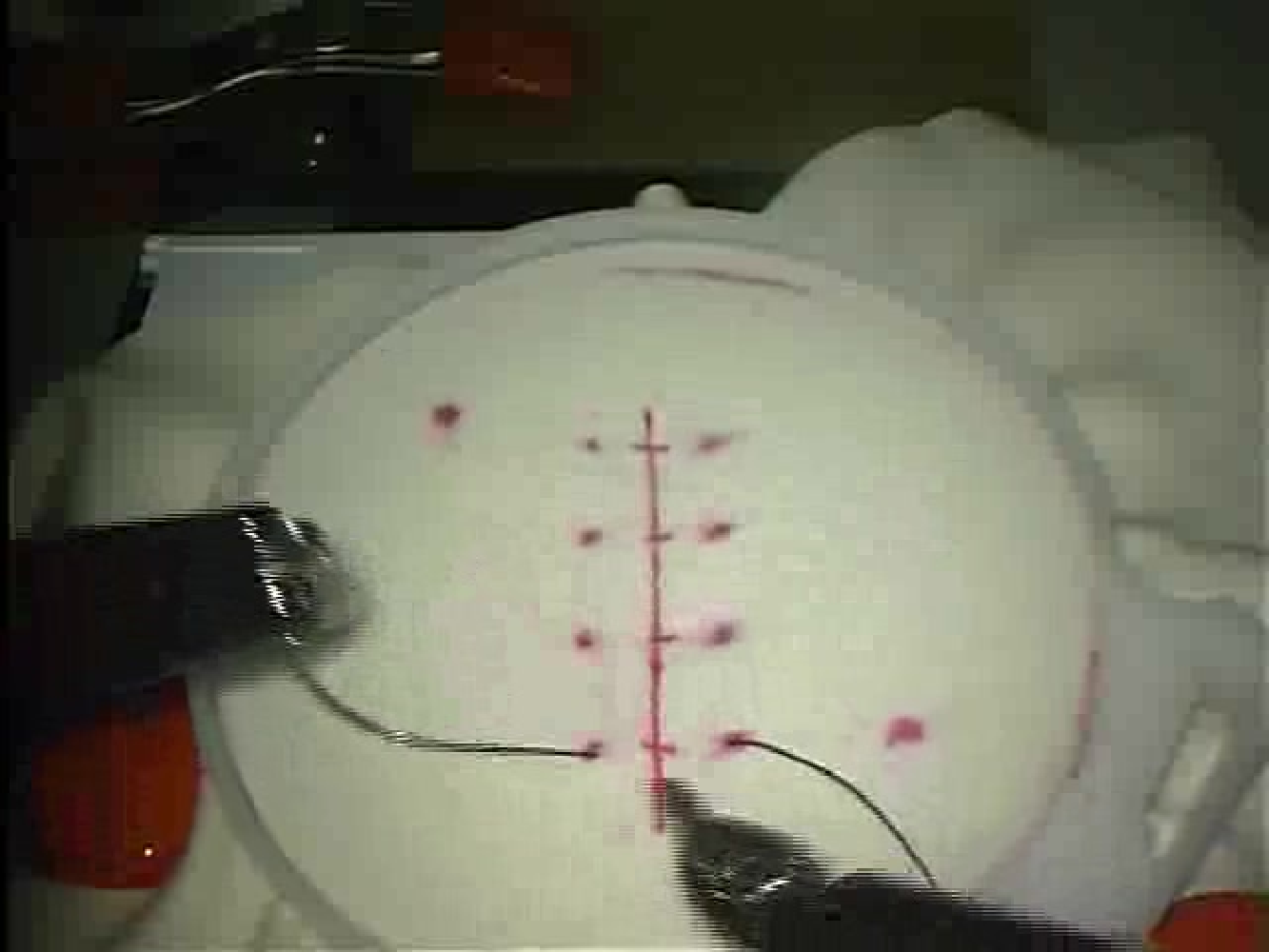}\includegraphics[width = 1in]{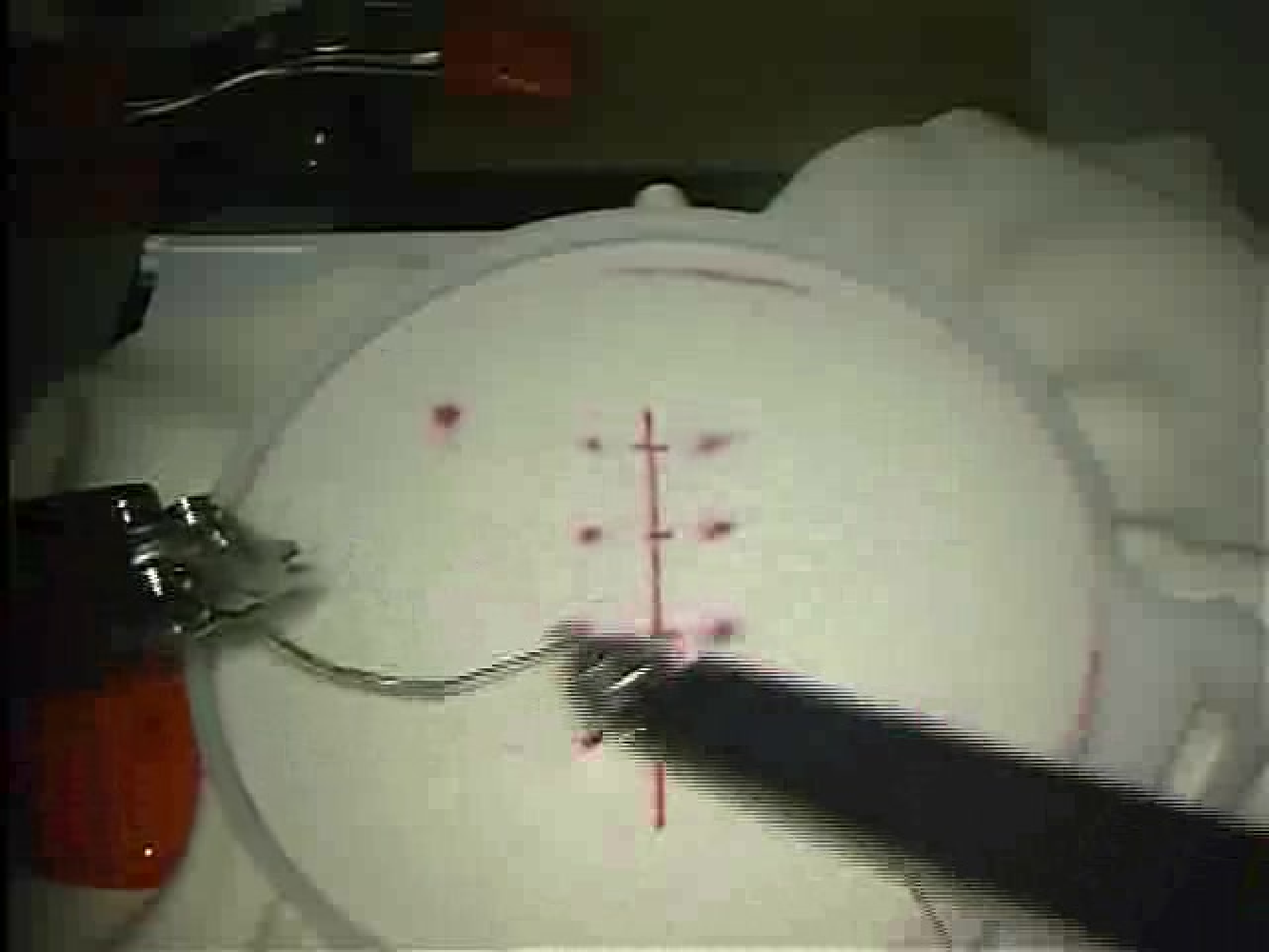}\includegraphics[width = 1in]{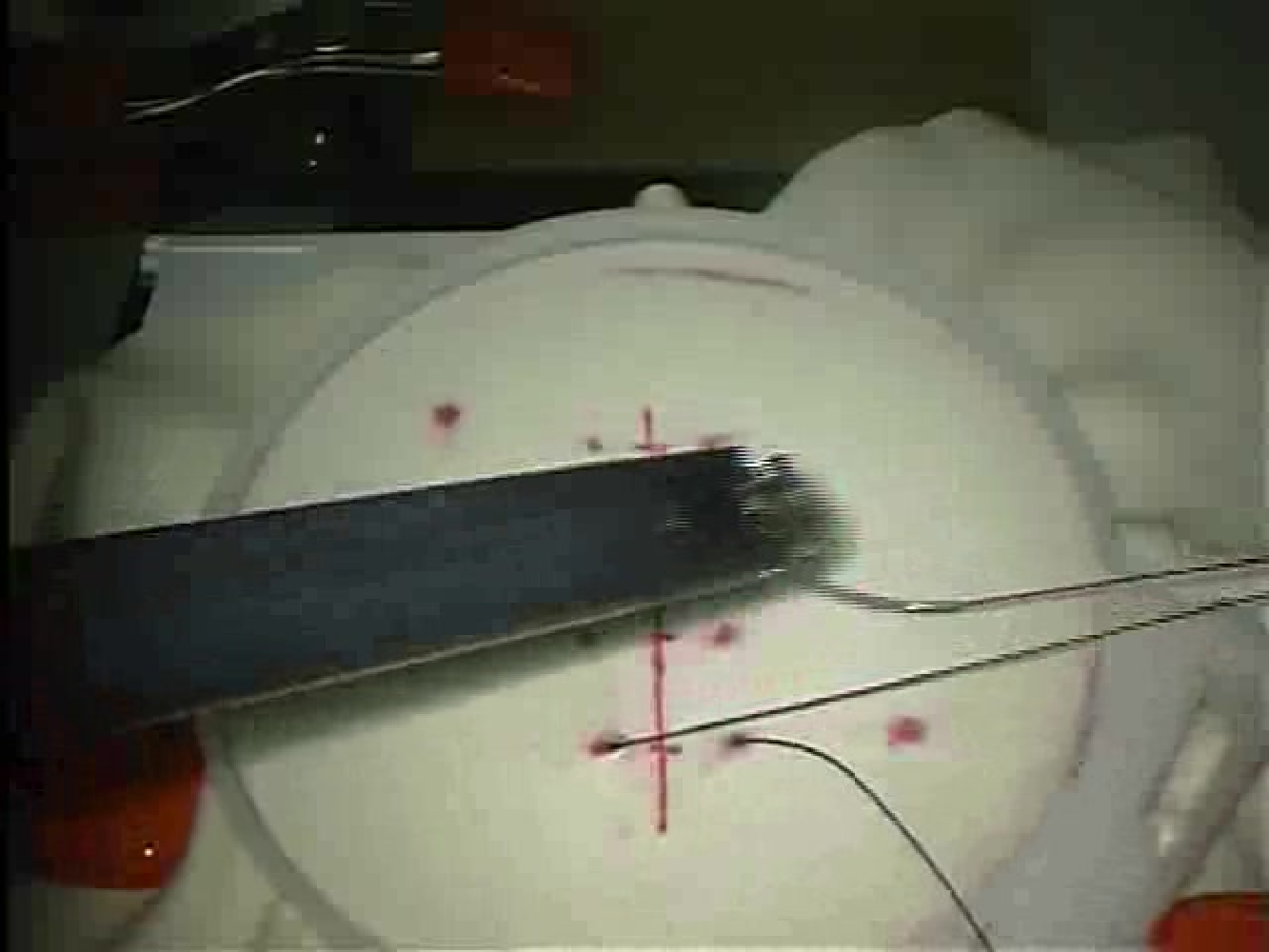}
\caption{ A temporal sequence of surgical activities during a \textit{Suturing} task are shown (from left to right, top to bottom). }
 \label{fig:suture}
\end{figure}

 The JHU-ISI Gesture and Skill Assessment Working Set (JIGSAWS) \cite{jigsaws}  provides a public benchmark surgical activity dataset. In this video dataset,  $8$ surgeons perform $3$ surgical tasks on the daVinci Surgical System (dVSS $^{\tiny{\textregistered}}$): \textit{Suturing} , \textit{Needle Passing} and \textit{Knot Tying}, and the dataset includes video data captured during the performance of these tasks from endoscopic cameras at $30$Hz. The dataset provides gesture labels $<G_{1},G_{2},...,G_{n}>$, which are the smallest action units such as \textit{Reaching for needle with right hand}. We performed our experiments on the \textit{Suturing} task which is composed of $10$ different gestures. A temporal sequence of surgical activities during a \textit{Suturing} task are shown in Figure \ref{fig:suture}.
 
\subsection{Annotation of Surgical Tool Poses} 

\begin{figure*}[!t]
\centering
\subfloat{\includegraphics[width = 2in]{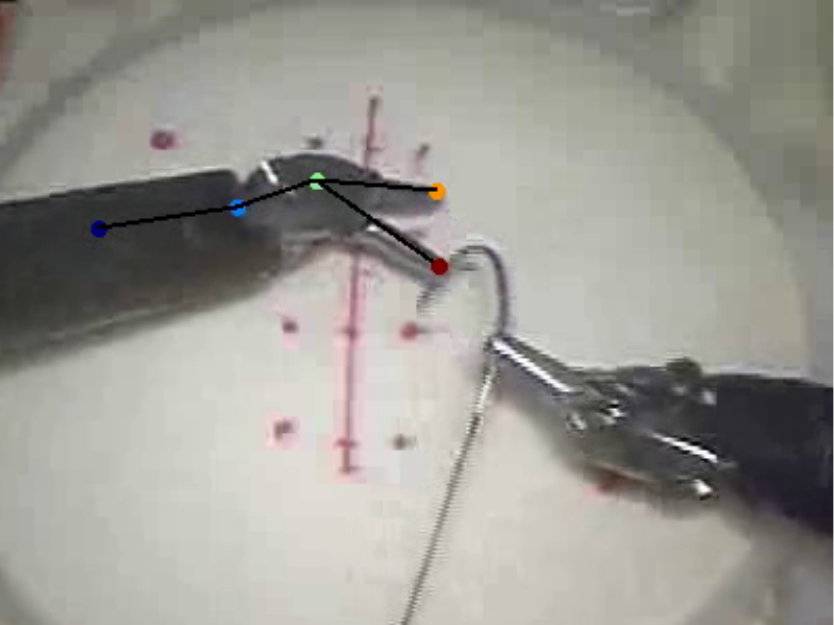}\includegraphics[width = 2in]{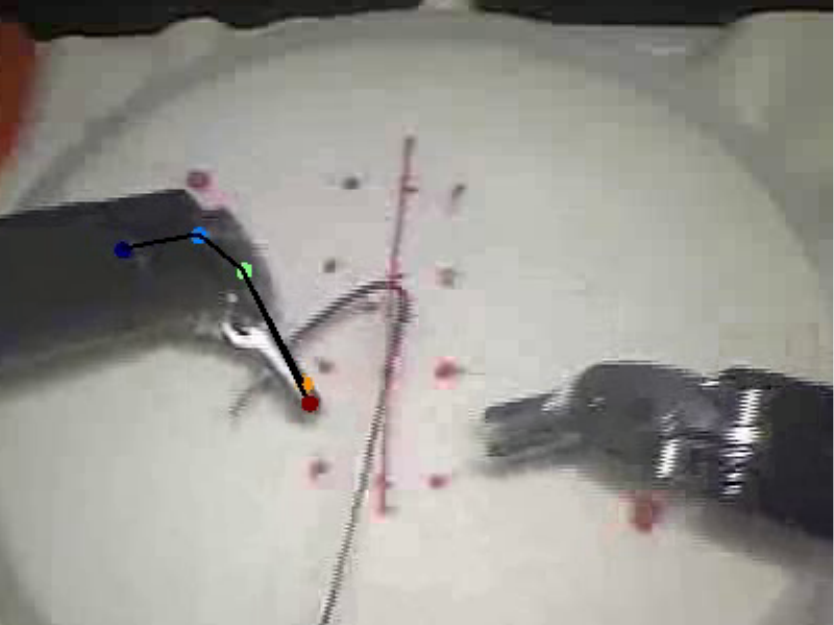}\includegraphics[width = 2in]{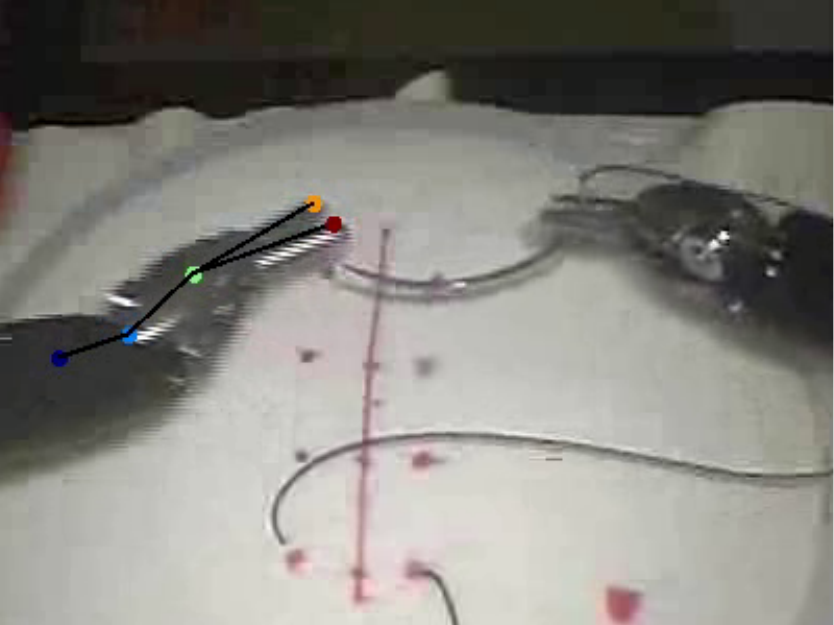}} 
\caption{Sample results of the pose estimation of the left tool are shown overlaid on the corresponding JIGSAWS video frames.}
 \label{fig:Suturing_labeled}
\end{figure*}

We labeled the surgical tool poses in a subset and then we trained a deep residual network (ResNet50) \cite{ResNet} to estimate the poses in the rest of the dataset. We first extracted frames, and then we clustered these frames using a simple $k-means$ clustering algorithm based on frame similarity.  We picked the $20$ most distinguishable frames based on these clusters in order to use for annotation. In other words, for each video we labeled only $20$ frames. Please note that, we intentionally used frames from the same videos in this step, for efficient labeling purposes with minimal effort (as pose estimation is not the focus of this work). We defined $5$ joints to capture the structure of a surgical grasper tool with respect to joints (the arm, the joint that connects the arm and the tool, the tool and its end effectors). We used DeepLabCut \cite{DeepLabCut} to both annotate the joints and to train the ResNet with transferred weights learned from ImageNet \cite{ImageNet}. Using the learned model, we estimated the pose coordinates for the rest of the frames. Sample results of the pose estimation of the left tool are shown overlaid on the corresponding JIGSAWS video frames in Figure \ref{fig:Suturing_labeled}. 

\subsection{Preprocessing videos as Input to ST-GCN}  \label{Input}

For each frame, we use the pose estimations and the confidence scores as input to construct our spatial temporal graph. We defined video segments $V_{t}  = (v_{1} ,v_{2},..., v_{t-1} , v_{t} )$ of $t=90$ consequent frames at $30$ fps which equals to $3$ seconds. We set the gesture label of this segment of activity as the gesture label of the frame at time $t=90$. We collected these video segments in a sliding window manner with a step size of $3$ frames, and we used these segments as input to ST-GCN. For the initial segments, we pad the frames to the beginning of the video segment by copying the first frame.

\section{Methods}

For each video segment, we construct an undirected spatial temporal graph as explain above \ref{Input} to form representations of the joints over time. A spatial configuration partitioning is then applied for constructing the convolution operations on graphs. The Spatial Temporal Graph Convolutional Network (ST-GCN) model is composed of multiple spatial temporal graph convolution operators (ST-GCN units), and applies multiple layers of spatial temporal convolutions on the neighbouring spatial and temporal nodes on the input graph. Using these convolutions, the hierarchical representations are learned which capture the spatial and temporal dynamics of surgical activities, instead of relying on rule based parsing techniques. Following multiple layers of graph convolutions and pooling, a soft-max layer is applied which gives the probability distribution for the corresponding surgical gesture labels. We explain each step of our methodology in detail below.

\subsection{Spatial Temporal Graph Construction}

For each video segment, we construct an undirected spatial temporal graph G = (V, E)  to represent the joints over temporal sequences of frames. First, for each frame, we define nodes corresponding each joint. We construct the spatial graphs by connecting these nodes with edges according to the connectivity of the surgical tool structure (skeleton). Then, for the temporal part, we connect each joint to the same joint in the consecutive frames forming inter-frame edges representing the trajectory of the joint over time.

\subsection{Spatial Temporal Graph Convolution Network (ST-GCN)}

After a spatial graph based on the joints of the surgical tool and the temporal edges between corresponding joints in consecutive frames are defined, a spatial configuration partitioning function is proposed for constructing the graph convolutional layer, which is then used to build the ST-GCN \cite{stgcn}. Considering the graph CNN model within one single frame at time $\tau$, where there are N joint nodes $V_{t}$ , along with the skeleton edges $E_{S}(\tau)=\left\{v_{t i} v_{t j} | t=\right.\tau,(i, j) \in H\}$, the input to the ST-GCN is the joint coordinate vectors on the graph nodes. Assuming an image as a regular 2D grid graph, a graph convolution operation can be applied. Dai et al. \cite{Dai2017}  proposes padding to the grid sampling locations on the image to enable free form deformation of the sampling grid. This way, the output feature map of convolution operation is also expressed as a 2D grid, and  the output feature maps can have the same size as the input feature maps. 

The 2D convolution consists of two steps \cite{Dai2017}: 1) sampling using a regular grid R over the input feature map and 2) a weighted summation step: summation of sampled values weighted by $w$.

Given a convolution operator with the kernel size of $K \times K$, and an input feature map $f_{i n}$ with the number of channels $c$, the output value for a single channel at the spatial location $x$ can be written as

\begin{equation} 
f_{o u t}(\mathbf{x})=\sum_{h=1}^{K} \sum_{w=1}^{K} f_{i n}(\mathbf{p}(\mathbf{x}, h, w)) \cdot \mathbf{w}(h, w)
\end{equation}

where the sampling function $\mathbf{p}: Z^{2} \times Z^{2} \rightarrow Z^{2}$ enumerates the neighbor pixels with respect
to the location $x$.

The weight function $\mathbf{w}: Z^{2} \rightarrow \mathbb{R}^{c}$ provides a weight vector in $c$-dimension real space for computing the inner product with the sampled input feature vectors of dimension $c$. Note that the weight function is irrelevant to the input location $x$. Standard convolution on the image domain is therefore achieved by encoding a rectangular grid in $p(x)$ \cite{stgcn}.

The convolution operation on a spatial graph is then defined by Yan et al. \cite{stgcn} by extending the formulation proposed by Dai et al. \cite{Dai2017}. Yan et al.  \cite{stgcn} redefine the sampling function $p$ and the weight function $w$ as explained below:

\subsubsection{Sampling function}

Extending the sampling function formulated by Dai et al. \cite{Dai2017}, Yan et al.  \cite{stgcn} define the  sampling function on the neighbor set $B\left(v_{t i}\right)=\left\{v_{t j} | d\left(v_{t j}, v_{t i}\right) \leq D\right\}$ of a node $v_{t i}$  where $d\left(v_{t j}, v_{t i}\right)$ denotes the minimum length of any path from $v_{t j}$ to $v_{t i}$. The sampling function $\mathbf{p}: B\left(v_{t i}\right) \rightarrow V$ can be written as 

\begin{equation} 
\mathbf{p}\left(v_{t i}, v_{t j}\right)=v_{t j}
\end{equation}

In this work, similarly to Yan et al.  \cite{stgcn}, we set the stride as $D = 1$, that is, the $1$-neighbor set of joint nodes.

\subsubsection{Weight function}

For graphs, where there is no inherent fixed spatial order (as there is in a 2D rigid graph referring an image), Yan et al.  \cite{stgcn} defines the order using a graph labeling process in the neighbor graph around the root node as proposed by Nieperth et al. \cite{Niepert2016}. They suggest partitioning the neighbor set $B\left(v_{t i}\right)$ of a joint node $v_{t i}$ into a fixed number of $K$ subsets, where each subset has a numeric label. Thus we can have a mapping $l_{t i}: B\left(v_{t i}\right) \rightarrow\{0, \ldots, K-1\}$ which maps a node in the neighborhood to its subset label.  The weight function $\mathbf{w}\left(v_{t i}, v_{t j}\right): B\left(v_{t i}\right) \rightarrow R^{c}$ can be implemented by  

\begin{equation}
\mathbf{w}\left(v_{t i}, v_{t j}\right)=\mathbf{w}^{\prime}\left(l_{t i}\left(v_{t j}\right)\right).
\end{equation}

\subsection{Partitioning}

We follow the partitioning strategy \textit{Spatial Configuration Partitioning} as proposed by  Yan et al.  \cite{stgcn} which utilizes the specific spatial configuration of the surgical tool skeleton. We use the strategy to divide the neighbor set into three subsets: the root node itself, the neighboring nodes that are closer to the gravity center of the skeleton than the root node (centripetal group) and the neighboring nodes that are further (centrifugal group). The gravity center is calculated as the average coordinate of all joints in the surgical tool skeleton at a single frame. This approach is then extended to the spatial-temporal domain.

We can formalize this partitioning strategy as below:

\begin{equation}
l_{t i} (v_{t} j)=\begin{cases}  
0 & \text { if } r_{j}=r_{i} \\ 1 & \text { if } r_{j}<r_{i} \\ 2 & \text { if } r_{j}>r_{i} 
\end{cases}
\end{equation}

where $r_{i}$ is the average distance from gravity center to joint $i$ over all frames in the training set.

\subsection{Spatial Graph Convolution}

Revisiting Eq. 1 with refined sampling function and weight function that are extended to graphs, we have 

\begin{equation}
f_{\text {out}}\left(v_{t i}\right)=\sum_{v_{t j} \in B\left(v_{t i}\right)} \frac{1}{Z_{t i}\left(v_{t j}\right)} f_{i n}\left(v_{t j}\right) \cdot \mathbf{w}\left(l_{t i}\left(v_{t j}\right)\right)
\end{equation}

where the normalizing term $ |Z_{t i}\left(v_{t j}\right)=\left|\quad\left\{v_{t k} | l_{t i}\left(v_{t k}\right)=\right.\right.\left.l_{t i}\left(v_{t j}\right)\right\}|  $ is added to balance the contributions of different subsets to the output and equals the cardinality of the corresponding subset.

\subsection{Spatial Temporal Modeling}

Now that we have the spatial graph convolution defined, we need to extend our model to temporal domain. We do this by extending the concept of neighbourhood to also include temporally connected joints as shown below \cite{stgcn}.

\begin{equation}
B\left(v_{t i}\right)=\left\{v_{q j}\left|d\left(v_{t j}, v_{t i}\right) \leq K,\right| q-t | \leq\lfloor\Gamma / 2\rfloor\right\}
\end{equation}

where $\Gamma$ is the temporal kernel size.

As proposed by Yan et al.  \cite{stgcn} , we keep the sampling function as the spatial one, and modify the weight function by extending the label map $l_{S T}$  for a spatial temporal neighborhood rooted at $v_{t i}$ to be

\begin{equation}
l_{S T}\left(v_{q j}\right)=l_{t i}\left(v_{t j}\right)+(q-t+\lfloor\Gamma / 2\rfloor) \times K
\end{equation}

where $l_{t i}\left(v_{t i}\right)$ is the label map for the single frame case at $v_{t i}$.

\subsection{Network Architecture}

\begin{figure*}[!t] 
\centering
 {\includegraphics[width = 4in]{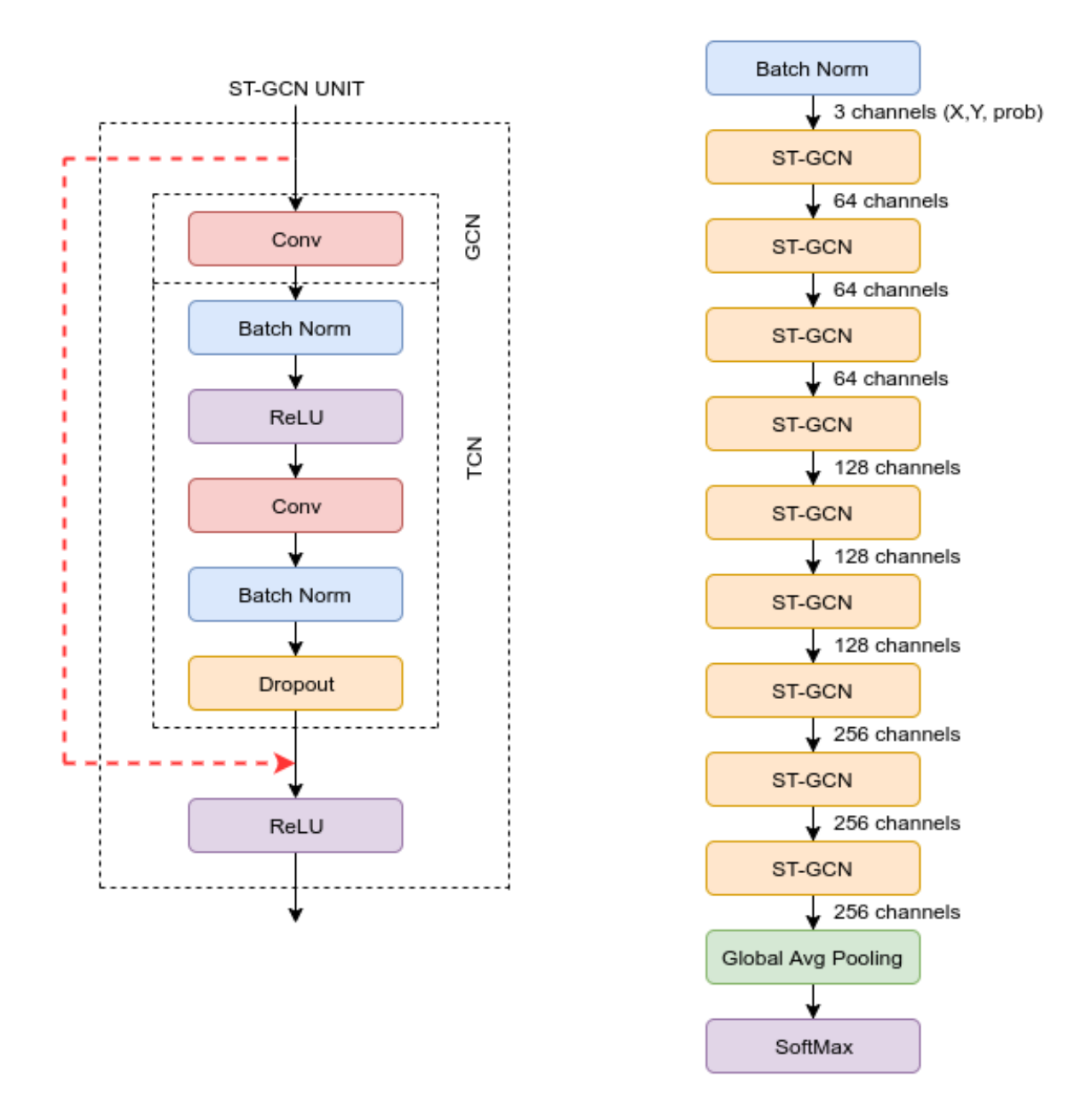}}
\caption{The ST-GCN model is composed of $9$ layers of spatial temporal graph convolution operators (ST-GCN units).  }
\label{fig:architecture}
\end{figure*}

The pose estimation and the adjacency matrix connecting the joints of the surgical tools is used as input to our ST-GCN, and then fed to a batch normalization layer. The spatial configuration partitioning function is used for constructing the graph convolutional layer, which is then used to build the ST-GCN \cite{stgcn}. The ST-GCN model is composed of $9$ layers of spatial temporal graph convolution operators (ST-GCN units), and applies multiple layers of spatial temporal convolutions on the neighbouring spatial and temporal nodes on the input graph. Using these convolutions, hierarchical representations which capture the spatial and temporal dynamics of surgical activities are learned. The first three layers of our architecture have $64$ channels, the following three layers have $128$ channels, and the last three layers have $256$ channels for output. These layers have $9$ temporal kernel size. The ResNet \cite{ResNet} mechanism is applied on each ST-GCN unit and a random dropout is performed. Following the mentioned multiple layers of graph convolutions and pooling, a $256$ dimension feature vector for each sequence is fed into a SoftMax classifier in order to label the sequences. Figure \ref{fig:architecture} demonstrates the architecture of the ST-GCN and the ST-GCN Unit.

\subsection{Training}
 
We trained the ST-GCN for $30$ epochs with stochastic gradient descent (SGD) optimization algorithm with a base learning rate of $0.01$ and then we decreased the learning rate using a step approach by diving the learning rate by $10$ at every $10$ epochs, we set the weight decay to $0.0005$. In order to avoid overfitting, we used a random dropout with $0.5$ probability. We also performed data augmentation; firstly, we performed random affine transformations which apply random combinations of different angle, translation and scaling factors on the skeleton sequences of all consequent frames. Secondly, we randomly sampled fragments from the skeleton sequences of consequent frames.

\subsection{Evaluation} 

We carried out our experiments with a TITAN X (Pascal architecture) GPU and an Intel Xeon (R) CPU E5 $3.50$ GHz$\times8$ with a $31.2$ GiB memory. All experiments are conducted on the PyTorch deep learning framework \cite{pytorch}.

For testing, we used the Leave-one-user-out (LOUO) experimentation split set which is provided by JIGSAWS. In the LOUO setup for cross-validation, there are eight folds, each one consisting of data from one of the eight subjects. We reported the average accuracy of all eight folds. We predicted the gesture label at every $3$ frames that is, $10$ frames per second. We compared the results of our model with the JIGSAWS Benchmark and the more recent CNN based studies (Table \ref{table:results}).

\textit{Suturing} task of JIGSAWS dataset has the chance baseline for gesture recognition of $10\%$ ( there are $10$ different gestures available). Our results demonstrate $68 (67.86) \%$ average accuracy on this dataset which suggests a significant improvement. Our experimental results show that learned spatial temporal graph representations of surgical videos are informative and they perform well in terms of recognizing low-level surgical activities (gestures) even when used individually.

\begin{table*}[]
 \centering
\resizebox{.55\textwidth}{!}{
\begin{tabular}{ll}
JIGSAWS Benchmark  \cite{jigsaws_benchmark}                                                                                                              & \begin{tabular}[c]{@{}l@{}}Average\\ Accuracy\end{tabular} \\ \hline
\multicolumn{1}{|l|}{GMM-HMM (kinematic)}                                             & \multicolumn{1}{l|}{73.95}                                 \\ \hline
\multicolumn{1}{|l|}{\begin{tabular}[c]{@{}l@{}}KSVD- SHMM (kinematic)\end{tabular}}          & \multicolumn{1}{l|}{73.45}                                 \\ \hline
\multicolumn{1}{|l|}{\begin{tabular}[c]{@{}l@{}}MsM-CRF (kinematic)\end{tabular}}           & \multicolumn{1}{l|}{67.84}                                 \\ \hline
\multicolumn{1}{|l|}{MsM-CRF (video)}                                                                       & \multicolumn{1}{l|}{77.29}                                 \\ \hline
\multicolumn{1}{|l|}{\begin{tabular}[c]{@{}l@{}}MsMCRF (kinematic + video)\end{tabular}}     & \multicolumn{1}{l|}{78.98}                                 \\ \hline
\multicolumn{1}{|l|}{\begin{tabular}[c]{@{}l@{}}SC-CRF (kinematic)\end{tabular}}              & \multicolumn{1}{l|}{81.74}                                 \\ \hline
\multicolumn{1}{|l|}{\begin{tabular}[c]{@{}l@{}}SC-CRF  (kinematic + video)\end{tabular}}      & \multicolumn{1}{l|}{81.60}                                 \\ \hline
CNN based models (Evaluation at 10 fps)                                                                         & \begin{tabular}[c]{@{}l@{}}Average\\ Accuracy\end{tabular} \\ \hline
\multicolumn{1}{|l|}{S-CNN (video) \cite{Colin2016}  }                                                                      & \multicolumn{1}{l|}{74.0}                                  \\ \hline
\multicolumn{1}{|l|}{ST-CNN (video) \cite{Colin2016}  }                                                               & \multicolumn{1}{l|}{77.7}                                  \\ \hline
\multicolumn{1}{|l|}{2D ResNet-18 (video) \cite{ResNet}   }                                                             & \multicolumn{1}{l|}{79.5}                                  \\ \hline
\multicolumn{1}{|l|}{3D CNN (K) + window (video) \cite{Funke2019} }                                                    & \multicolumn{1}{l|}{84.3}                                  \\ \hline
                                             ST-GCN (Evaluation at 10 fps)                                                                              & \begin{tabular}[c]{@{}l@{}}Average\\ Accuracy\end{tabular} \\ \hline
\multicolumn{1}{|l|}{Ours (2D joint pose estimations (X,Y coordinates) from video)}                   & \multicolumn{1}{l|}{67.86}                                   \\ \hline
\end{tabular}}
\caption{We compared the results of our model with the JIGSAWS Benchmark and the recent CNN based studies. The tests are performed on the Suturing task of the JIGSAWS dataset, Leave-one-user-out (LOUO) experimentation split. We reported the average accuracy of all eight folds of this setup. } 

\label{table:results}
\end{table*}

\section{Conclusion}

Modeling and recognition of surgical activities poses an interesting research problem as the need for assistance and guidance through automation is addressed by the community. Although a number of recent works studied automatic recognition of surgical activities, generalizability of these works across different tasks and different datasets remains a challenge. In order to overcome the challenge of generalizability across different tasks and different datasets, we need to define generic and sparse representations of surgical activities that are robust to scene variation. Pose-based joint and skeleton representations suffer relatively little from the intra-class variances when compared to image cues \cite{yao}. These representations are also able to infer part information of objects such as orientational and relative spatial relationships.

In this paper, we introduced a modality independent of the scene, therefore robust to scene variation, based on spatial temporal graph representations of surgical tools. To our knowledge, our paper is the first to use pose-based skeleton representations for surgical activity recognition. To explore the effectiveness of the modality we introduce, we modeled and recognized surgical activities in videos using this modality. We first constructed a spatial graph of surgical tool joints representing the surgical tool skeleton, we then extend this graph temporally. We learn hierarchical spatial temporal features using ST-GCN \cite{stgcn} instead of relying on rule based parsing techniques. ST-GCN exploits the natural graph structure of skeleton data and the structural connectivities of joints over time across a sequence of frames.

Our experimental results show that learned spatial temporal graph representations of surgical videos are informative and they perform well in terms of recognizing low-level surgical gestures even when used individually. We experiment our model on the \textit{Suturing} task of the JIGSAWS dataset where the chance baseline for gesture recognition is $10\%$ ( there are $10$ different gestures available). Our results demonstrate $68\%$ average accuracy on this dataset which suggests a significant improvement, and suggests the generic representations learned are meaningful for surgical activity recognition. These learned representations can be used either individually, in cascades or as a complementary modality in surgical activity recognition, therefore provide a benchmark for future studies. 

A limitation of our work is the dependency on the joint information, and for this study we opted for an efficient labeling/training set up with minimal effort as pose estimation is not the focus of this work. However, pose estimation of surgical tools is a well studied problem and recent works \cite{Du2018} have reached an accuracy above 96\% (average of all joints) even for in-vivo datasets (Endovis (https://github.com/surgical-vision/EndoVisPoseAnnotation)). There is also room for improvement, in terms of applying a temporal smoothing algorithm for post-processing the predictions to filter out flicker of predictions. 

Our ST-GCN implementation will be made publicly available upon acception (subject to approval of the funding body). Please see ``Supplementary Files'' for sample results (video).

\section*{Acknowledgment}
This work was supported by French state funds managed within the Investissements d'Avenir program by BPI France (project CONDOR). All experiments are conducted on the PyTorch deep learning framework \cite{pytorch}, we built our implementation on the publicly released source of the work by Yan et al. \cite{stgcn}. We used DeepLabCut \cite{DeepLabCut} to both annotate the joints and to train the ResNet \cite{ResNet} with transferred weights learned from ImageNet \cite{ImageNet}.

\end{document}